\documentclass[11pt,a4paper]{article}
\usepackage[hyperref]{emnlp-ijcnlp-2019}
\usepackage{times}
\usepackage{latexsym}
\usepackage{url}
\usepackage{tabularx}
\usepackage{graphicx}
\usepackage[linesnumbered,boxed,ruled,commentsnumbered]{algorithm2e}
\usepackage{amsmath}
\usepackage{amsfonts}
\usepackage{paralist}
\usepackage{xcolor}
\usepackage{booktabs}

\aclfinalcopy 

\newcommand{\modelname}{\textbf{KBRD}}
\newcommand{\modeltitle}{\textbf{K}nowledge-\textbf{B}ased \textbf{R}ecommender \textbf{D}ialog System}
\newcommand{\redial}{R{\small E}D{\small IAL}}

\title{Towards Knowledge-Based Recommender Dialog System}

\author{
    Qibin Chen$^{1}$, Junyang Lin$^{2}$, Yichang Zhang$^{2}$, Ming Ding$^{1}$, Yukuo Cen$^{1}$, Hongxia Yang$^{2}$, Jie Tang$^{1}$\\
    $^{1}$Department of Computer Science and Technology, Tsinghua University \\
    $^{2}$DAMO Academy, Alibaba Group \\
    \texttt{\{cqb19,dm18,cyk18\}@mails.tsinghua.edu.cn} \\
    \texttt{\{junyang.ljy,yichang.zyc,yang.yhx\}@alibaba-inc.com}\\
    \texttt{jietang@tsinghua.edu.cn}
}

\date{}

\begin{document}
\maketitle
\begin{abstract}
    In this paper, we propose a novel end-to-end framework called \modelname, which stands for \modeltitle. It integrates the recommender system and the dialog generation system.
    The dialog system can enhance the performance of the recommendation system by introducing knowledge-grounded information about users' preferences, and the recommender system can improve that of the dialog generation system by providing recommendation-aware vocabulary bias. 
    Experimental results demonstrate that our proposed model has significant advantages over the baselines in both the evaluation of dialog generation and recommendation. 
    A series of analyses show that the two systems can bring mutual benefits to each other, and the introduced knowledge 
    contributes to both their performances.\footnote{Code will be available at \url{https://github.com/THUDM/KBRD}.}

\end{abstract}

\section{Introduction}

Dialog in e-commerce has great commercial potential.
In conventional recommender systems, personalized recommendation is highly based on the previous actions of users, including searching, clicking and purchasing. 
These actions can be regarded as users' feedbacks that reflect users' interest. However, due to its implicitness, such feedback can only reflect a part of users' interest, causing inaccuracy in recommendation.
Another information source about user preferences is the dialog between users and services. In such dialog, users often provide more information about their preferences. They often ask for tips or recommendation in the dialog. In this process, services can guide them to speak out their interests in order to solve users' problems and meet their requirements. Compared with the implicit feedback, the feedback from the dialog is more explicit and more related to users' preferences. Therefore, a recommender dialog system possesses high commercial potential.

\begin{table}[tb]
    \small
	\centering
	\begin{tabularx}{\linewidth}{lX}
		\toprule
		\textbf{USER}: & Hello!  \\
		\textbf{RECOMMENDER}: & What kind of movies do you like? \\
		\textbf{USER}: & I am looking for a movie recommendation. When I was younger \\ & I really enjoyed the \emph{A Nightmare on Elm Street (1984)}.  \\
		\midrule
		\textbf{BASELINE}: & Have you seen \emph{It (2017)}? \\
		\midrule
		\textbf{OURS}:  & I like horror movies too! Have your seen \emph{Halloween (1978)} ? \\
		\midrule
		\textbf{HUMAN}: & Oh, you like scary movies? I recently watched \emph{Happy Death Day (2017)}. \\
		\bottomrule
	\end{tabularx}
	\caption{\textbf{An example of the outputs of recommender dialog systems.} The model recommends items (italic) while maintaining the dialog with the user. Compared with the baseline, our dialog system gives a more diverse and consistent response.}
	\label{tab-human}
\end{table}

A recommender dialog system can be regarded as a combination of a recommender system and a dialog system. 
A dialog system should respond to users' utterances with informative natural language expressions, and a recommender system should provide high-quality recommendation based on the content of users' utterances. 
We demonstrate an example in Table~\ref{tab-human}. 
In brief, a recommender dialog system should perform well in both tasks. 

An ideal recommender dialog system is an end-to-end framework that can effectively integrate the two systems so that they can bring mutual benefits to one another. 
In this setting, information from the recommender system can provide vital information to maintain multi-turn dialog, while information from the dialog system that contains implication of users' preferences can enhance the quality of recommendation. 
Besides, the incorporation of external knowledge can 
strengthen the connections between systems and enhance their performances.
Therefore, driven by the motivations, we propose a novel end-to-end framework that integrates the two systems.
We name it \modelname, standing for \modeltitle. 


\begin{figure*}[tb]
    \centering
    \includegraphics[width=\linewidth]{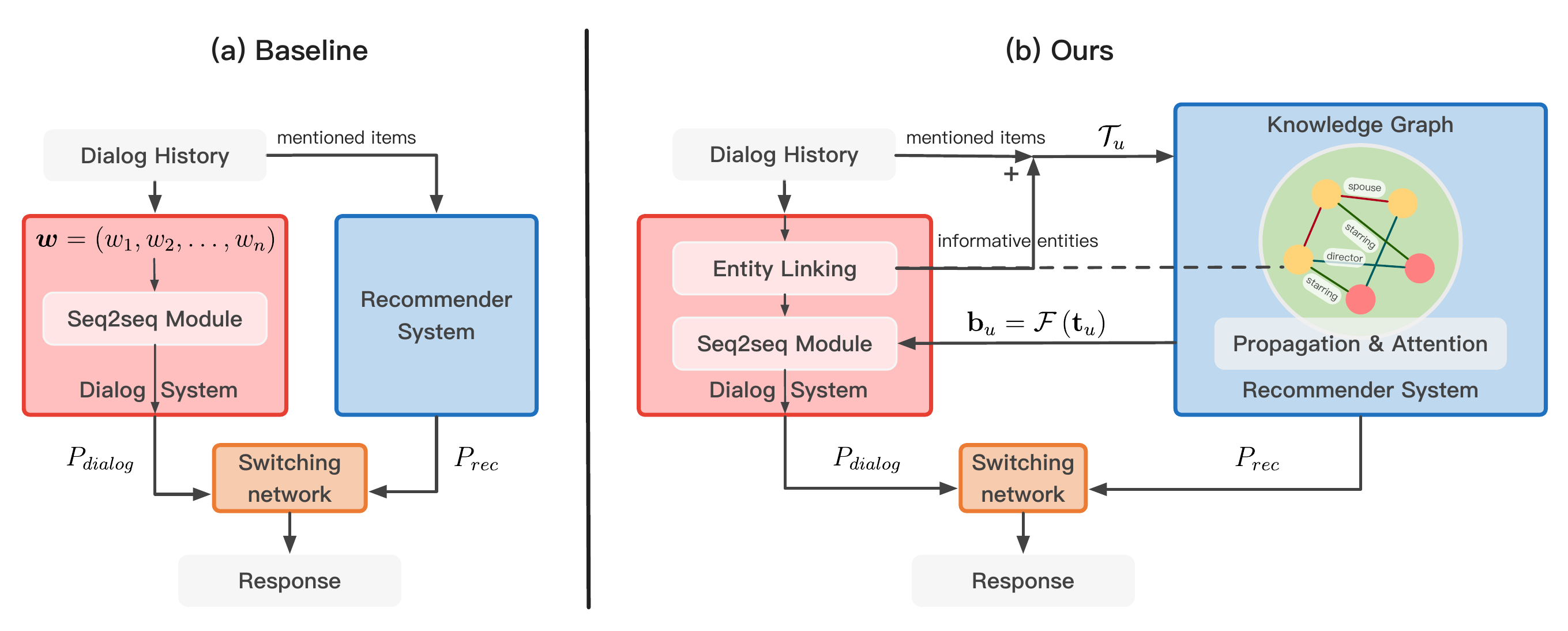}
    \caption{
    \textbf{Comparative illustration on modules of the existing baseline framework and our proposed \modelname\ framework.}
    (a) The connection between the recommender system and the dialog system in the baseline framework is 
    weak.
    The dialog system takes the plain text of the dialog history as input and the recommender only considers mentioned items in the dialog.
    (b) Our framework enables interaction between the two systems.
    First, informative entities are linked to an external knowledge graph and sent to the recommender besides items.
    They are propagated on the KG via a relational graph convolutional network, enriching the representation of user interest.
    Second, the knowledge-enhanced user representation is sent back to the dialog system in the form of vocabulary bias, enabling it to generate responses that are consistent with the user's interest.
    }
    \label{fig:overview}
\end{figure*}

Specifically, the dialog generation system provides contextual information about items to the recommender system. 
For instance, for a movie recommendation system, contextual information can be director, actor/actress and 
genre.
Thus, even with no item mentioned in the dialog, the recommender system can still perform high-quality recommendation based on the contextual information. 
In return, the recommender system provides recommendation information to promote the dialog, such as recommendation-aware vocabulary bias. 
Furthermore, we incorporate external knowledge into our framework. The knowledge graph helps bridge the gap between systems and enhances both their performances.

We conduct a series of experiments that demonstrate the effects of our framework in both the evaluation of recommendation and dialog generation. Moreover, the analyses show that dialog information effectively tackles the cold-start problem in recommendation, and the recommendation-aware vocabulary bias from the recommender system improves the quality of generated dialogs. Also, the biased words can be parts of reasons to explain the system's decisions for recommendation.

\section{Preliminary}
Before we introduce our proposed framework, we provide an illustration of the basic framework of the recommender dialog system to show how the recommendation system and the dialog system are organized for end-to-end training.

\subsection{Recommender System}
Provided with a user's information, a recommender system is aimed at retrieving a subset of items that meet the user's interest from all the items.
In a cold-start setting, 
the recommender system initially has no knowledge about the user.
With the progress of the dialog, 
the recommender system accumulates user's information and builds a user profile.
Thus, it can provide reasonable recommendation based on the user preferences reflected in the conversation. 


To implement an effective recommender system in this task, it is available to build a recommender system based on conventional collaborative filtering algorithms \citep{Sarwar:2001WWW} or based on neural networks \citep{he2017neural}. For example, \citet{li_towards_2018}
applies a user-based autoencoder \citep{sedhain_autorec:_2015} to recommend new items based on previously mentioned items
in the dialog.

\subsection{Dialog System}
The dialog system in the basic framework is in charge of generating multi-turn dialog with a natural language generation model. 
The pioneering work~\cite{li_towards_2018} on conversational recommendation task adopted Hierarchical Recurrent Encoder Decoder (HRED)~\citep{sordoni_conv_2015, hred, hred_dialog} for this part.
The HRED is an encoder-decoder framework for sequence-to-sequence learning \citep{seq2seq}.
In the framework, an encoder receives the dialog history as input and encodes it to high-level representation, while a decoder generates responses based on the encoded representation.
By recursively encoding and decoding the previous information, the system makes utterances in the multi-turn dialog.


\subsection{End-to-End System}
In order to perform end-to-end training, we demonstrate the combination of the recommender system and conversation system.
Specifically, the input of the recommender system is constructed based on the dialog history, which is a representation of mentioned items in the dialog.
The output of the recommender system $P_{\mathrm{rec}}$, which is a probability distribution over the item set, can be combined with the output of the dialog system $P_{\mathrm{dialog}} \in \mathbb{R}^{|V|}$, where $V$ refers to the vocabulary. 
A switching mechanism \citep{gulcehre2016pointing} controls the decoder to decide whether it should generate a word from the vocabulary or an item from the recommender output at a certain timestep.
\begin{align}
    P(w)&=p_{s} P_{\mathrm{dialog}}(w) + \left(1-p_{s}\right) P_{\mathrm{rec}}(w)
\end{align}
\begin{align}
    p_{s}&=\sigma\left(w_{s}o+b_{s}\right)
\end{align}
where $w$ represents either a word from the vocabulary or an item from the item set, 
$o$ is the hidden representation in the final layer of the dialog system.
$w_s \in \mathbb{R}^{d}$ and $b_s \in \mathbb{R}$ are the switcher's parameters and $\sigma$ refers to the sigmoid function.
Therefore, the whole system can be trained in an end-to-end fashion.

\section{Proposed Model}
In this section, we introduce our proposed framework \modelname\ that integrates the recommender system and the dialog system effectively via knowledge propagation.
We show how knowledge connects the two systems and how they bring mutual benefits to each other.

\subsection{Dialog-Aware Recommendation with Knowledge}
Recommendation of the basic framework is solely based on the mentioned items in the dialog history.
Such recommendation ignores contextual information in dialog  
that often indicates users' preferences.

Here we propose to make use of the dialog contents, including the non-item information, in the process of recommendation.
Furthermore, to effectively recommend items from the non-item information, we introduce an external knowledge graph from DBpedia \citep{lehmann2015dbpedia} to our system.
The knowledge can build a connection between dialog contents and items.

\paragraph{Incorporating Dialog Contents}
Specifically, we have a knowledge graph $\mathcal{G}$ consisting of triples $(h, r, t)$ where $h, t \in \mathcal{E}$ and $r \in \mathcal{R}$.
$\mathcal{E}$ and $\mathcal{R}$ denote the sets of entities and relations in the knowledge graph.
We first match each item in the item set to entities in $\mathcal{E}$ by name.\footnote{For example, a movie item ``star wars'' is matched to \url{http://dbpedia.org/resource/Star_Wars_(film)} in the dbpedia knowledge graph.}
We then perform entity linking \citep{daiber2013improving} on dialog contents
and thus informative non-item entities appearing in dialog contents are matched to $\mathcal{E}$.\footnote{Utterance ``I like science fiction movies.'' is associated with \url{http://dbpedia.org/resource/Science_fiction_film}. An utterance may be associated with one or multiple entities.}
Therefore, we can represent a user as $\mathcal{T}_u = \left\{e_1, e_2, \cdots, e_{|\mathcal{T}_u|}\right\}$, where $e_i \in \mathcal{E}$.
To be more specific, 
it is a set of mentioned items plus non-item entities extracted from the dialog contents, linked to the knowledge graph. 

\paragraph{Relational Graph Propagation}
Inspired by \citet{schlichtkrull2018modeling},
we apply Relational Graph Convolutional Networks (R-GCNs) to encode structural and relational information in the knowledge graph to entity hidden representations.
An intuition behind this is that neighboring nodes in knowledge graph may share similar features that are useful for recommendation.
For example, when a user speaks of his/her preference on an actor/actress, the recommender should provide movies that have a close connection to that person.
In addition, by taking different relations into consideration, the system models different types of neighbors more accurately.

Formally, at layer $0$, we have a trainable embedding matrix $\mathbf{H}^{(0)} \in \mathbb{R}^{|\mathcal{E}| \times d^{(0)}}$ for nodes (i.e., entities) on the knowledge graph.
Then, for each node $v$ in $\mathcal{E}$ at layer $l$, we compute:
\begin{align*}
h_{v}^{(l+1)}=\sigma\left(\sum_{r \in \mathcal{R}} \sum_{w \in \mathcal{N}_{v}^{r}} \frac{1}{c_{v, r}} W_{r}^{(l)} h_{w}^{(l)}+W_{0}^{(l)} h_{v}^{(l)}\right)
\label{eq:rgcn}
\end{align*}
where $h_{v}^{(l)} \in \mathbb{R}^{d^{(l)}}$ denotes the hidden representation of node $v$ at the $l$-th layer of the graph neural network, and $d^{(l)}$ denotes the dimensionality of the representation at the layer.
$\mathcal{N}_{v}^{r}$ denotes the set of neighbor indices of node $ v $  under relation $r \in \mathcal{R}$.
$W_{r}^{l}$ is a learnable relation-specific transformation matrix for vectors from neighboring nodes with relation $r$.
$W_{0}^{l}$ is a learnable matrix for transforming the nodes' representation at the current layer.
$c_{v, r}$ is a normalization constant that can either be learned or chosen in advance (e.g., $c_{v, r}=\left|\mathcal{N}_{v}^{r}\right|$).

For each node on the graph, it receives and aggregates the messages from its neighboring nodes after relation-specific transformation.
Then it combines the information with its hidden representation to form its updated representation at the next layer.

Finally, at the last layer $L$, structural and relational information is encoded into the entity representation $h_{v}^{(L)}$ for each $v \in \mathcal{E}$.
We denote the resulting knowledge-enhanced hidden representation matrix for entities in $\mathcal{E}$ as $\mathbf{H}^{(L)} \in \mathbb{R}^{|\mathcal{E}|\times d^{(L)}}$.
We omit the $(L)$ in the following paragraphs for simplicity.

\paragraph{Entity Attention}
The next step is to recommend items to users based on knowledge-enhanced entity representations. 
While an item corresponds to an entity on the knowledge graph, a user may have interacted with multiple entities.
Given $\mathcal{T}_u$,
we first look up the knowledge-enhanced representation of entities in $\mathcal{T}_u$ from $\mathbf{H}$, and we have:
\begin{align}
\mathbf{H}_u = (h_1, h_2, \cdots, h_{|\mathcal{T}_u|}) 
\end{align}
where $h_i \in \mathbb{R}^{d}$ is the hidden vector 
of entity $e_i$.
Here our objective is to 
encode
this vector set of variable size to a vector of fixed size so that we can compute the similarity between user and item.
Instead of simply averaging these vectors, we choose a linear combination of the $|\mathcal{T}_u|$ vectors.
Specifically, we apply self-attention mechanism \citep{lin+al-2017-embed-iclr} that takes $\mathbf{H}_u$ as input and outputs a distribution $\alpha_{u}$ over $|\mathcal{T}_u|$ vectors:
\begin{align}
\alpha_u=\operatorname{softmax}\left(w_{a 2} \tanh \left(W_{a 1} \mathbf{H}_{u}^{T}\right)\right)
\end{align}
where $W_{a1} \in \mathbb{R}^{d_{a}\times d}$ is a weight matrix and $w_{a2} $ is a vector of parameters with size $d_{a}$. The final representation of user $u$ is computed as follows:
\begin{align}
    t_u = \alpha_u\mathbf{H}_{u}
    \label{eq:user_final_hidden}
\end{align}

This enables the recommender system to consider the importance of different items and non-item entities in the dialog.
Finally, the output of our recommender is computed as follows:
\begin{align}
P_{\mathrm{rec}} = \operatorname{softmax}(\operatorname{mask}(t_{u}\mathbf{H}^{T}))
\end{align}
where $\operatorname{mask}$ is an operation that sets the score of non-item entities to $-\infty$. The masking operation ensures that the recommendations are all items. 

\subsection{Recommendation-Aware Dialog}
\label{sec-rec_aware_dialog}

Instead of applying HRED, we introduce the Transformer framework to the dialog system in this task. 
Transformer \citep{transformer} can reach significantly better performances in many tasks, such as machine translation \citep{transformer,ott_transformer}, question answering \citep{rajpurkar2016squad,yang2018hotpotqa,ding2019cognitive} and natural language generation \citep{wiki_sum,KOBE}. 
In our preliminary experiments, we have found that Transformer can also achieve better performance than HRED in this task, and thus we apply this framework to the dialog system.

The Transformer is also an encoder-decoder framework for sequence-to-sequence learning. 
The Transformer encoder consists of an embedding layer and multiple encoder layers. 
Each encoder layer has a self-attention module and a Point-Wise Feed-Forward Network (FFN). The encoder encodes the dialog history $\boldsymbol{x} = (x_1, x_2, \ldots, x_n)$ to high-level representations $\boldsymbol{s} = (s_1, s_2, \ldots, s_n)$. 
Similarly, the Transformer decoder contains an embedding layer and multiple decoder layers with self-attention and FFN. Moreover, each of them contains a multi-head context attention to extract information from the source-side context. The decoder generates a representation $o$ at each decoding time step.



In order to predict a word at each decoding time step, the top layer of the decoder, namely the 
output
layer, generates a probability distribution over the vocabulary:
\begin{align}
P_{\mathrm{dialog}}=\operatorname{softmax}\left(Wo + b\right)
\label{fig:vocab}
\end{align}
where $W \in \mathbb{R}^{|V| \times d}$ and $b \in \mathbb{R}^{|V|}$ are weight and bias parameters, and $V$ refers to the vocabulary.

However, so far the dialog system is completely conditioned on the plain text of the dialog contents.
By further introducing the recommender system's knowledge of the items that have appeared in dialog, we guide the dialog system to generate responses that are more consistent with the user's interests.
Specifically, we add a vocabulary bias $b_u$ to the top layer of the decoder inspired by \citet{michel-neubig-2018-extreme}.
Different from their work, $b_u$ is computed based on the recommender system's hidden representation of user $u$:
\begin{align}
    b_u = \mathcal{F}(t_u)
\end{align}
where $\mathcal{F}: \mathbb{R}^{d} \to \mathbb{R}^{|V|}$ represents a feed-forward neural network and $t_{u}$ is the user representation in the recommendation context introduced in Equation~\ref{eq:user_final_hidden}.

Therefore, the computation of the top layer of the decoder becomes:
\begin{align}
P_{\mathrm{dialog}}=\operatorname{softmax}\left(Wo + b + b_{u} \right)
\end{align}

So far, we have built an end-to-end framework that bridges the recommender system and the dialog system, which enables mutual benefits between the systems.

\section{Experiments}

In this section, we provide an introduction to the details of our experiments, including dataset, setting, evaluation as well as further analyses.

\subsection{Dataset}
\textit{REcommendations through DIALog (\redial)} is a dataset for conversational recommendation.
\citet{li_towards_2018} collected the dialog data and built the dataset through Amazon Mechanical Turk (AMT). With enough instructions, the workers on the platform generated dialogs for recommendation on movies.
Furthermore, in order to achieve and dialog-aware recommendation, besides movies, we introduce the relevant entities,  such as director and style, from DBpedia. The number of conversations is 10006 and the number of utterances is 182150. The total number of users and movies are 956 and 51699 respectively. 


\subsection{Setting}
We implement the models in PyTorch and train on an NVIDIA 2080Ti.
For the recommender, both the entity embedding size $d^{(0)}$ and the hidden representation size $d^{(l)}$ are set to 128. We choose the number of R-GCN layers $L=1$ and the normalization constant $c_{v,r}$ to 1.
For Transformer, all input embedding dimensions and hidden sizes are set to 300.
During training, the batch size is set to 64.
We use Adam optimizer \citep{adam} with the setting $\beta_{1}=0.9$, $\beta_{2}=0.999$ and $\epsilon=1\times10^{-8}$.
The learning rate is $0.003$ for the recommender and $0.001$ for the Transformer.
Gradient clipping restricts the norm of the gradients within [0, 0.1].


\subsection{Evaluation Metrics}

The evaluation of dialog consists of automatic evaluation and human evaluation. The metrics for automatic evaluation are perplexity and distinct n-gram. Perplexity is a measurement for the fluency of natural language. Lower perplexity refers to higher fluency. Distinct n-gram is a measurement for the diversity of natural language. Specifically, we use distinct 3-gram and 4-gram at the sentence level to evaluate the diversity. As to human evaluation, we collect ten annotators with knowledge in linguistics and require them to score the candidates on the consistency with the dialog history. We sample 100 multi-turn dialogs from the test set together with the models' corresponding responses, and require them to score the consistency of the responses.\footnote{Note that we did not provide the utterances of the baseline Transformer to annotators. Based on our observation, the generations of the Transformer-based models are significantly different from those of \redial. In case that annotators had knowledge about models, we did not require them to score the utterances of Transformer.} The range of score is 1 to 3.

The evaluation for recommendation is Recall@K.
We evaluate that whether the top-k items selected by the recommender system contain the ground truth recommendation provided by human recommenders.
Specifically, we use Recall@1, Recall@10, and Recall@50 for the evaluation.

\begin{table}[tb]
	\centering
	\begin{tabular}{lrrr}
		\toprule
		Model & R@1 & R@10 & R@50\\
		\midrule
		\redial & 2.3$\pm$0.2 & 12.9$\pm$0.7 & 28.7$\pm$0.9 \\
		\midrule
		\modelname\ \textbf{(D)} & 2.7$\pm$0.2 & 14.0$\pm$0.6 & 30.6$\pm$0.7 \\
		\modelname\ \textbf{(K)} & 2.6$\pm$0.2 & 14.4$\pm$0.9& 31.0$\pm$1.2 \\
		\modelname & \textbf{3.0}$\pm$0.2 & \textbf{16.3}$\pm$0.3 & \textbf{33.8}$\pm$0.7 \\
		\bottomrule
	\end{tabular}
	\caption{\textbf{Evaluation of the recommender system.} We report the results of Recall@1, Recall@10 and Recall@50 of the models ($p \ll 0.01$). \modelname{} \textbf{(D)} stands for only incorporating the dialog contents. \modelname{} \textbf{(K)} stands for only incorporating knowledge. The results demonstrate that both the interaction with the dialog system and the external knowledge are helpful for the improvement of model performance, and our proposed model reaches the best performance on the three metrics.}
	\label{tab-rec}
\end{table}

\subsection{Baselines}
The baseline models for the experiments are illustrated in the following:
\begin{itemize}
    \item \textbf{\redial} This is a basic model for conversational recommendation. It basically consists of a dialog generation system based on HRED \citep{hred, hred_dialog}, a recommendation system based on autoencoder and a sentiment analysis module.
    \item \textbf{Transformer} We name our implemented baseline model Transformer. It is similar to \redial, but its dialog generation system is based on the model Transformer \citep{transformer}. Except for that, the others remain the same.
\end{itemize}

\subsection{Results}
\label{sec-result}

In the following, we present the results of our experiments, including the model performances in recommendation and dialog generation. 

\paragraph{Recommendation}
To evaluate the effects of our recommendation system, we conduct an evaluation of Recall@K. We present the results in Table~\ref{tab-rec}. From the results, it can be found that our proposed model reaches the best performances in the evaluation of Recall@1, Recall@10 and Recall@50. Furthermore, we also demonstrate an ablation study to observe the contribution of the dialog system and the introduced knowledge. It can be found that either dialog or knowledge can bring improvement to the performance of the recommendation system. Their combination improves the performance the most by +0.7 Recall@1, +3.4 Recall@10 and +5.1 Recall@50, which are advantages of 30.4\%, 26.4\% and 17.8\% respectively. This shows that the information from both sources is contributive. The dialog contains users' preferred items as well as attributes, such as movie director and movie style, so that the system can find recommendation based on these inputs. The knowledge contains important features of the movie items so that the system can find items with similar features.
Further, the combination brings an advantage even greater than sum of the two parts, which proves the effectiveness of our model.

\begin{table}[tb]
	\centering
	\begin{tabular}{lrrrr}
		\toprule
		Model & PPL & Dist-3 & Dist-4 & CSTC \\
		\midrule
		\redial & 28.1 & 0.11 & 0.13 & 1.73 \\
		Transformer & 18.0 & 0.27 & 0.39 & - \\
		\modelname & \textbf{17.9} & \textbf{0.30} & \textbf{0.45} & \textbf{1.99} \\
		\bottomrule
	\end{tabular}
	\caption{\textbf{Automatic and human evaluation of dialog generation.}
	For automatic evaluation, we evaluate the perplexity (PPL) and distinct n-gram (Dist-3 and Dist-4 refer to distinct 3-gram and 4-gram respectively) of the generated dialogs. 
	For human evaluation, we ask human annotators to evaluate the consistency (CSTC) of the generated utterances with the dialog history. 
	Our proposed method performs the best in all evaluations compared with the baselines.}
	\label{tab-dialog}
\end{table}

\paragraph{Dialog}
Table~\ref{tab-dialog} shows the results of the evaluation of the baseline models and our proposed method in dialog generation. 
In the evaluation of perplexity, Transformer has much lower perplexity (18.0) compared to \redial\ (28.1),
and \modelname\ can reach the best performance in perplexity. 
This demonstrates the power of Transformer in modeling natural language. 
In the evaluation of diversity, we find that the models based on Transformer significantly outperform \redial\ from the results of distinct 3-gram and 4-gram. 
Besides, it can be found that \modelname\ has a clear advantage in diversity over the baseline Transformer. 
This shows that our model can generate more diverse contents without decreasing fluency. 

As to the human evaluation, we ask human annotators to score the utterances' consistency with their dialog history. Compared with \redial\, \modelname\ reaches better performance by +0.22 consistency score, which is an advantage of 15\%. Moreover, considering the range is between 1 and 3, this is a large gap between the performances of the two models in this evaluation. 
To make a consistent response in a dialog, the model should understand the dialog history and better learn the user's preference. The baseline \redial\ does not have a strong connection between the dialog system and user representation. Instead, in our framework, the recommender system provides the recommendation-aware vocabulary bias $b_u$, which is based on the user representation $t_u$, to the dialog system. Thus the dialog system gains knowledge about the user's preference and generates a consistent response.

\section{Discussion}


In this section, we conduct a series of analyses to observe the effects of our proposed model. We discuss 
how
dialog can improve the recommendation performance and 
how
recommendation can enhance the dialog quality. 

\begin{figure}[tb]
    \centering
    \includegraphics[width=\linewidth]{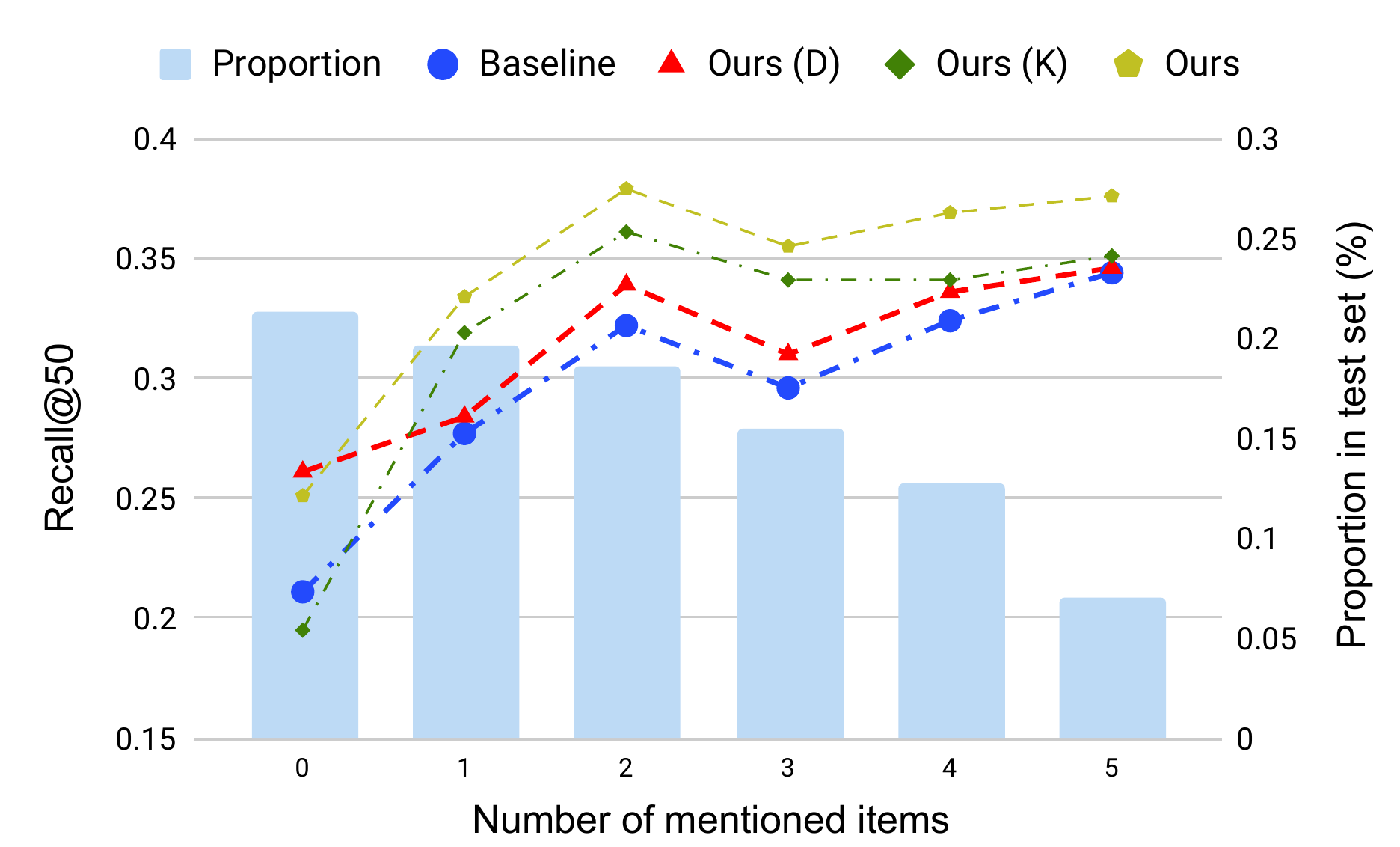}
    \caption{\textbf{Performance of the recommender system with different numbers of mentioned items.} The x-axis refers to the number of mentioned items in the dialog, the y-axis for the line chart (on the left) refers to the model performance on the Recall@50 evaluation, and the y-axis for the histogram (on the right) refers to proportion in the test set.
    This shows recommendation is much more difficult with few items mentioned (i.e., at the first few rounds in dialog).
    Leveraging dialog contents makes a great difference in this situation.}
    \label{fig:turns}
\end{figure}

\begin{table*}
    \small
	\centering
	\begin{tabular}{lcccccccc}
		\toprule
		Movie & 1 & 2 & 3 & 4 & 5 & 6 & 7 & 8\\
		\midrule
		Star Wars & space & alien & sci-fi & star & sci & robot & smith & harry \\
		The Shining & creepy & stephen & gory & horror & scary & psychological & haunted & thriller \\
		The Avengers (2012) & marvel & superhero & super & dc & wait & batman & thor & take \\
		Beauty and the Beast & cute & disney & animated & live & music & child & robin & kids \\
		\bottomrule
	\end{tabular}
	\caption{\textbf{Examples of top 8 vocabulary bias}.
	Given a mentioned movie, we visualize words with high probability based on the recommendation-aware vocabulary bias.
	The examples show that the biased words have strong connections with their corresponding movies.
	While the vocabulary bias benefits the dialog system to generate responses that are more consistent with user interest, they also contribute to the explainability of recommendation.}
	\label{tab-bias}
\end{table*}

\subsection{Does dialog help recommendation?}
We first evaluate whether the dialog contents can benefit the recommendation system.
The results of the evaluation are demonstrated in Figure~\ref{fig:turns}.
From the histogram in the figure,
we observe that most of the dialogs contain only a few mentioned movies.
The dialogs with only 0-2 mentioned movies take up a proportion of 62.8\% of the whole testing dataset.
Therefore, it is important for the system to perform high-quality recommendation with only a small number of mentioned movies. This also corresponds to the classical problem ``cold start''
\citep{schein2002methods}
in the recommender system.
In real applications, we also expect that the system can perform high-quality recommendation with fewer rounds.
This represents the efficiency of the recommender system, which can save users' time and efforts.

Specifically, we demonstrate the performances of four systems in Figure~\ref{fig:turns}.
They are the basic framework, the one only with the interaction with the dialog system, the one only with the external knowledge and \modelname\ with both dialog and knowledge incorporation. 
From the figure, it can be found that while there is no mentioned item in the dialog, the baseline and the one only with knowledge perform the worst. 
In contrast, the two models with dialog incorporation perform significantly better. 
This shows that the context in the dialog contains much useful non-item information about users' preferences, such as director, actor/actress in movie recommendation.
Therefore, while there is no mentioned item, the recommender system can still perform high-quality recommendation based on the contextual information. 
With the increase of mentioned items, the contribution of knowledge becomes more significant than the dialog. 
On average, the system with both information sources performs the best.
Dialog introduces contextual information and knowledge introduces movie features and structural connection with other movies.

\subsection{Does recommendation help dialog?}
In Section~\ref{sec-result}, we present the performances of the baselines and our model \modelname\ in dialog generation. It can be found that the interaction with the recommendation system can enhance the performance of the dialog system in both automatic evaluation and human evaluation. Also, an example of the responses of different models is shown in Table~\ref{tab-human}. With the dialog history, the baseline \redial\ simply uses a generic response with a recommended movie. Instead, \modelname\ has more concern about the mentioned items apart from the plain text of dialog history.
The user representation from our recommender system contains such information, which is sent to the dialog system to form a vocabulary bias.
With such information, \modelname\ has a better understanding of both the dialog history as well as the user's preference, and thus generates a consistent response.

To further study the effects of the recommender system on dialog generation, we display the top biased words from the vocabulary bias.
Note that in \modelname\ a connection between the recommender system and dialog system is the recommendation-aware vocabulary bias $b_u$.
To be specific, we compute the recommendation-aware bias $b_u$ in dialog and select the components with the top-8 largest values.\footnote{After stop words filtering.}
Then we record the corresponding words and observe whether these words are related to the mentioned movies. 
We present several examples in Table~\ref{tab-bias}.
From the table, we observe that the words are highly related to the mentioned movies. For example, when ``The Shining'' is mentioned, some of the top biased words are ``creepy'', ``gory'' and ``scary'', which are consistent with the style of the horror movie, and ``stephen'', who is the original creator of the movie.
Therefore, it can be suggested that the recommendation system conveys important information to the dialog system in the form of a vocabulary bias.
Furthermore, these biased words can also serve as 
explicit explanation to
recommendation results. 
From this perspective, this shows the interpretability of our 
model.


\section{Related Work}

Recommender systems aim to find a small set of items that meet users' interest based on users' historical interactions.
Traditional recommender systems rely on collaborative filtering \citep{resnick1994grouplens, Sarwar:2001WWW}, and recent advances in this field rely much on neural networks \citep{Wang:2015KDD, he2017neural, ying2018graph}.
To deal with the cold-start problem and the sparsity of user-item interactions which these methods usually suffer,
researchers have proposed methods to incorporate external information, such as heterogeneous information networks \citep{yu2014personalized}, knowledge bases \citep{zhang2016collaborative, wang2018ripplenet} and social networks \citep{DBLP:conf/recsys/JamaliE10}.
Besides accuracy, explainability is also an important aspect when evaluating recommender systems \citep{zhang2014explicit, zhang2018explainable, wang2018reinforcement}. 

End-to-end dialog systems based on neural networks have shown promising performance in open-ended settings \citep{vinyals2015neural, sordoni_conv_2015, dodge_evaluating_2015, wen2015semantically} and goal-oriented applications \citep{bordes_learning_2016}.
Recent literature also explores the intersection of end-to-end dialog systems with other intelligence systems and creates new tasks such as visual dialog \citep{das2017visual, de2017guesswhat}, conversational recommendation \citep{li_towards_2018}.
In particular, \citet{li_towards_2018} collects a dataset of conversations focused on providing movie recommendations and proposes a baseline model for end-to-end training of recommender and dialog systems.
Earlier studies in this field focus on different tasks such as minimizing the number of user queries \citep{christakopoulou2016towards}, training the dialog agent to ask for facet values for recommendation \citep{sun2018conversational}.
Related literature can also be found in \citet{thompson2004personalized}, \citet{mahmood2009improving}, \citet{chen2012critiquing}, \citet{widyantoro2014framework} and \citet{li2019deep}.






\section{Conclusion}

In this paper, we propose a novel end-to-end framework, \modelname, which bridges the gap between the recommender system and the dialog system via knowledge propagation. 
Through a series of experiments, we show that \modelname\ can reach better performances in both recommendation and dialog generation in comparison with the baselines.
We also discuss how the two systems benefit each other. 
Dialog information is effective for the recommender system especially in the setting of cold start, and the introduction of knowledge can strengthen the recommendation performance significantly. 
Information from the recommender system that contains the user preference and the relevant knowledge can enhance the consistency and diversity of the generated dialogs.

\section*{Acknowledgements}
The work is supported by 
NSFC for Distinguished Young Scholar (61825602),
NSFC (61836013), and a research fund supported by Alibaba.
The authors would like to thank Lei Li and Chang Zhou for their insightful feedback, and responsible reviewers of EMNLP-IJCNLP 2019 for their valuable suggestions. Jie Tang is the corresponding author.

\bibliography{emnlp-ijcnlp-2019}
\bibliographystyle{acl_natbib}


\end{document}